\documentclass{article}

\usepackage{arxiv}

\usepackage[utf8]{inputenc} 
\usepackage[T1]{fontenc}    
\usepackage{hyperref}       
\usepackage{url}            
\usepackage{booktabs}       
\usepackage{amsfonts}       
\usepackage{nicefrac}       
\usepackage{microtype}      
\usepackage{lipsum}

\usepackage{bm}
\usepackage{amsmath}
\usepackage{amssymb}
\usepackage{amsfonts}
\usepackage{graphicx}
\def\vec#1{\mathbf{#1}}
\usepackage{algorithm}
\usepackage{algorithmic}
\usepackage{comment}

\title{Active Learning for Regression with Aggregated Outputs}

\author{
  Tomoharu Iwata\\
  NTT Communication Science Laboratories
}
\date{}

\begin{document}
\maketitle
\begin{abstract}
Due to the privacy protection or the difficulty of data collection, we cannot observe individual outputs for each instance, but we can observe aggregated outputs that are summed over multiple instances in a set in some real-world applications. To reduce the labeling cost for training regression models for such aggregated data, we propose an active learning method that sequentially selects sets to be labeled to improve the predictive performance with fewer labeled sets. For the selection measurement, the proposed method uses the mutual information, which quantifies the reduction of the uncertainty of the model parameters by observing the aggregated output. With Bayesian linear basis functions for modeling outputs given an input, which include approximated Gaussian processes and neural networks, we can efficiently calculate the mutual information in a closed form.  With the experiments using various datasets, we demonstrate that the proposed method achieves better predictive performance with fewer labeled sets than existing methods.
\end{abstract}

\section{Introduction}

Data are often aggregated for privacy protection, cost reduction,
or the difficulty of data collection~\cite{li2012compressed,armstrong1999geographically,bhowmik2019learning}.
For example, census data are averaged over spatial regions,
IoT data are aggregated to reduce the communication overhead,
the gene expression level is measured for each set of multiple cells,
and brain imaging data are observed for each set of voxels.
Since learning from such aggregated data
is important for applications where only aggregated data are available,
many machine learning methods for aggregated data have been
proposed~\cite{musicant2007supervised,quadrianto2009estimating,bhowmik2015generalized}.

Although 
the predictive performance of the machine learning model improves
as the number of training labeled data increases in general,
obtaining many labeled data incurs considerable cost.
Active learning has been successfully used for reducing the labeling cost,
where instances to be labeled are sequentially selected to improve the predictive performance~\cite{settles2009active,liang2018investigating,tang2019self,sugiyama2009pool}.
However, there have been no active learning methods for regression with aggregated data.

In this paper, we propose an active learning method for regression with aggregated outputs.
At the beginning of the active learning process,
we are given unlabeled sets of instances.
Then, for each active learning step,
we select a set to observe its aggregated output,
where we cannot observe outputs for each instance.
Our aim is to improve the predictive performance
of the outputs for each test instance.
The proposed method selects a set that maximizes the mutual information between
the aggregated output and model parameters,
which corresponds to the reduction of the uncertainty of the model parameters
by observing the aggregated output of the set.
Mutual information-based active learning has been successfully
used for non-aggregated data~\cite{mackay1992information,lawrence2003fast,krishnapuram2004semi}.

\begin{figure}[t]
\centering
\includegraphics[width=23em]{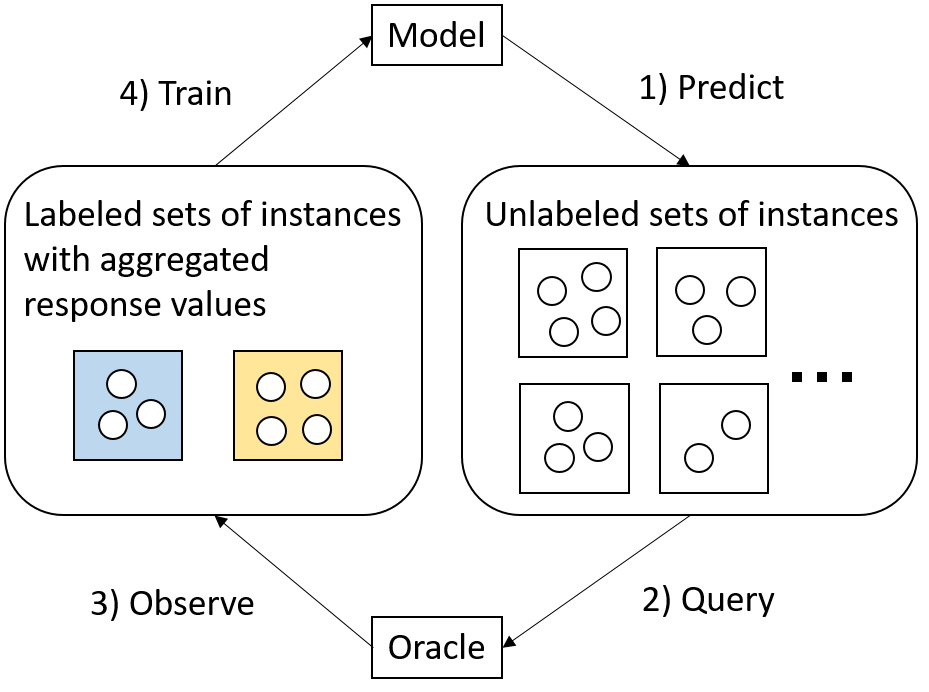}
\caption{Our framework of active learning with aggregated outputs. In the beginning, we are given unlabeled sets of instances. For each step, we iterate the following procedures: 1) Predict the distribution of the aggregated output for each of the unlabeled sets using the model. 2) Select a set from the unlabeled sets to be labeled using mutual information calculated based on the predicted distributions, and query the oracle. 3) Observe the aggregated output of the selected set, include it in the labeled sets, and exclude it from the unlabeled sets. 4) Retrain the model using the updated labeled sets.}
\label{fig:framework}
\end{figure}

We derive the mutual information using
linear basis function models as a regression model
that predicts the non-aggregated output given an input vector.
Various regression models can be formulated
as a linear basis function model, which include polynomial regression,
approximated Gaussian processes~\cite{rahimi2007random},
and neural networks by changing basis functions.
With the Bayesian inference framework of the linear basis function models,
we can model the distribution of the aggregated output
as a Gaussian distribution, and
we can calculate the mutual information on aggregated outputs
efficiently in a closed form.
Figure~\ref{fig:framework} shows the framework of our active learning.

The major contributions of this paper are as follows:
\begin{enumerate}
\item We propose the first active learning method for regression with aggregated outputs.
\item The proposed method is based on entropy and mutual information,
which are calculated efficiently using Bayesian linear basis function models,
and considers the correlation among the instances in each set.
\item We demonstrate the effectiveness of the proposed method with various datasets
compared with existing active learning methods for non-aggregated data.
\end{enumerate}
The remainder of this paper is organized as follows.
In Section~\ref{sec:related}, we briefly review related work.
In Section~\ref{sec:preliminaries}, we describe the probability distribution
of the weighted sum of Gaussian distributed random variables,
which is used in the proposed method.
In Section~\ref{sec:proposed}, we define our task, and propose
our active learning method for regression with aggregated outputs based on entropy and mutual information.
In Section~\ref{sec:experiments}, we evaluate the performance of our method by comparing existing methods.
Finally, we present concluding remarks and a discussion of future work
in Section~\ref{sec:conclusion}.

\section{Related work}
\label{sec:related}

Several frameworks of learning from aggregated data have been
proposed~\cite{goodman1953ecological,freedman1991ecological,zhang2020learning}.
Learning from label proportions~\cite{quadrianto2009estimating,patrini2014almost}
considers classification tasks, where outputs are categorical.
Multiple instance learning~\cite{maron1998framework} learns classification models from labeled sets,
where each set is positively labeled if at least one individual is positive, and it is otherwise negatively labeled.
Collective graphical models learn from contingency tables~\cite{sheldon2013approximate,kumar2013collective,iwata2020co}.
Regression from aggregated data has also been considered, where outputs are
continuous~\cite{bhowmik2015generalized,yousefi2019multi,tanaka2019spatially,law2018variational}.
Summed or averaged values are assumed to be observed in~\cite{park2014ludia,yousefi2019multi,tanaka2019spatially,law2018variational}
as with our setting,
and histograms are assumed to be observed in~\cite{bhowmik2015generalized}.
Some methods assume that both inputs and outputs are aggregated~\cite{bhowmik2015generalized},
and others assume that only outputs are aggregated while inputs
are not aggregated~\cite{musicant2007supervised,kuck2005learning}.
In this paper, we consider regression with aggregated outputs by a linear weighted summation.

Many active learning methods have been
proposed~\cite{settles2009active,tong2001support,bordes2005fast,bachman2017learning,liang2018investigating,tang2019self},
which include those for multiple instance learning~\cite{carbonneau2018bag},
and those for learning from label proportions~\cite{poyiadzis2019active}.
However, they are not for regression with aggregated outputs, and they are inapplicable to our task.
Batch active learning~\cite{kirsch2019batchbald,guo2007discriminative,pinsler2019bayesian}
selects multiple instances to be labeled, where
outputs for each instance are observed.
It is different from our task, where aggregated outputs are observed,
but individual outputs cannot be observed.

\section{Preliminaries}
\label{sec:preliminaries}

Let $\vec{y}=[y_{1},\cdots,y_{N}]\in\mathbb{R}^{N}$ be
jointly Gaussian distributed random variables,
\begin{align}
\vec{y}\sim\mathcal{N}(\vec{y}|\bm{\mu},\bm{\Sigma}),
\end{align}
where $\mathcal{N}(\cdot|\bm{\mu},\bm{\Sigma})$ is
the Gaussian distribution with mean $\bm{\mu}\in\mathbb{R}^{N}$
and covariance $\bm{\Sigma}\in\mathbb{R}^{N\times N}$.
The weighted sum of the random variables
$\bar{y}=\sum_{n=1}^{N}\theta_{n}y_{n}=\bm{\theta}^{\top}\vec{y}\in\mathbb{R}$
follows the Gaussian
distribution~\cite{johnson2002applied,lemons2002introduction},
\begin{align}
\bar{y}\sim\mathcal{N}(\bar{y}|\bm{\theta}^{\top}\bm{\mu},\bm{\theta}^{\top}\bm{\Sigma}\bm{\theta}),
\label{eq:sum_of_g}
\end{align}
where $\theta_{n}\in\mathbb{R}$ is the weight, and
$\bm{\theta}=[\theta_{1},\cdots,\theta_{N}]\in\mathbb{R}^{N}$.

\section{Proposed method}
\label{sec:proposed}

In Section~\ref{sec:task},
we define our task of active learning for aggregated outputs.
In Section~\ref{sec:model},
we present our model for predicting outputs
that are trained from 
labeled sets with aggregated outputs based on linear basis function models.
In Sections~\ref{sec:entropy} and \ref{sec:mutual},
we propose entropy-based and mutual information-based active learning
methods using our model that select a set to be observed next
to improve the predictive performance, respectively.
In Section~\ref{sec:procedures},
we present the procedures of the proposed method.

\subsection{Problem formulation}
\label{sec:task}

Suppose that we are given sets of input vectors
$\{\vec{X}_{a}\}_{a=1}^{A}$,
where $\vec{X}_{a}=\{\vec{x}_{an}\}_{n=1}^{N_{a}}$
is the $a$th set of input vectors,
$\vec{x}_{an}\in\mathbb{R}^{D}$ is the $n$th input vector,
$D$ is the number of attributes,
and $N_{a}$ is the number of input vectors in the set.
For each active learning step,
we select a set from $\{1,\cdots,A\}$, and
observe the aggregated output of selected set $a$
that is obtained by the weighted sum
of the output of the input vectors in the set,
\begin{align}
\bar{y}_{a}=\sum_{n=1}^{N_{a}}\theta_{an}y_{an}=\bm{\theta}_{a}^{\top}\vec{y}_{a}\in\mathbb{R},
\label{eq:bary}
\end{align}
where $\bar{\cdot}$ indicates that it is an aggregated value,
$y_{an}\in\mathbb{R}$
is the unknown output of input vector $\vec{x}_{an}$,
$\vec{y}_{a}=[y_{a1},\cdots,y_{aN_{a}}]\in\mathbb{R}^{N_{a}}$,
and
$\bm{\theta}_{a}=[\theta_{a1},\cdots,\theta_{aN_{a}}]\in\mathbb{R}^{N_{a}}$
is the weights.
We assume that weights $\bm{\theta}_{a}$ for all sets are known.
For example, 
$\theta_{an}=1$ when the aggregated data are obtained by summation,
and $\theta_{an}=\frac{1}{N_{a}}$ when they are obtained by average.
Our aim is to improve the test predictive performance of
the outputs with as few observations of aggregated outputs
as possible.
Table~\ref{tab:notation} shows our notation.
Although we assume that outputs are scalar,
the proposed method can be straightforwardly extended to
multivariate outputs.
When the aggregated response value is obtained by
integral $\bar{y}_{a}=\int\theta_{an}y_{an}dn$,
we can apply the proposed method
by approximating the integral by the summation
by dividing the space into a finite number of bins.

\begin{table}[t!]
  \centering
\caption{Notation.}
\label{tab:notation}
\begin{tabular}{ll}
\hline
Symbol & Description\\
\hline
$\vec{x}_{an}$ & input vector of the $n$th instance of the $a$th set.\\
$\vec{X}_{a}$ & $a$th set of input vectors.\\
$y_{an}$ & unobserved output value of the $n$th instance
of the $a$th set.\\
$\bar{y}_{a}$ & aggregated output value of the instances
in the $a$th set.\\
$\bm{\theta}_{a}$ & linear weights for aggregation of the $a$th set.\\
$N_{a}$ & number of instances in the $a$th set.\\
$D$ & number of attributes.\\
$\phi(\cdot)$ & basis function.\\
$\vec{w}$ & linear projection vector of the linear basis
function model, or parameters of the neural
network model.\\
$\mathcal{D}$ & labeled sets with aggregated outputs.\\
\hline
\end{tabular}
\end{table}

\subsection{Model}
\label{sec:model}

Let $\hat{y}(\vec{x};\vec{w})$ be a regression model
to predict the non-aggregated output given input vector $\vec{x}$,
where $\vec{w}$ is the parameters to be considered as random variables.
We consider the following linear basis function model,
\begin{align}
  \hat{y}(\vec{x};\vec{w})=\vec{w}^{\top}\phi(\vec{x}),
  \label{eq:linear}
\end{align}
where $\phi(\vec{x})\in\mathbb{R}^{K}$ is the nonlinear basis function
that transforms a $D$-dimensional input vector
to a $K$-dimensional vector, and $\vec{w}\in\mathbb{R}^{K}$.
A wide variety of regression models can be formulated by a linear basis function model,
which include linear regression, polynomial regression, approximated Gaussian processes with random features-based basis functions~\cite{rahimi2007random},
and neural networks with the last layer represented by random variables and neural network-based basis functions.
For aggregated data, linear regression~\cite{smith2014poverty,bhowmik2015generalized,wang2016crime},
Gaussian processes~\cite{tanaka2019refining,flaxman2015supported,law2018variational,tanaka2019spatially},
and neural networks~\cite{rahimi2007random} have been used.
The linear basis function formulation is preferable especially when training data are small,
which is a situation of active learning,
since the number of random variable parameters to be estimated is small,
and the posterior of the model parameters are estimated analytically.

\subsection{Entropy-based active learning}
\label{sec:entropy}

In the non-aggregated setting,
the entropy-based active learning selects an instance that maximizes the entropy of the output~\cite{sebastiani2000maximum}.
It corresponds to the uncertainty sampling~\cite{lewis1994sequential,settles2009active},
which selects a set whose aggregated output is least certain,
where the uncertainty is quantified by the entropy. The high entropy indicates the high uncertainty.
In the aggregated setting,
a set that maximizes the entropy of aggregated ouput $\bar{y}_{a}$
given input vector $\vec{X}_{a}$ 
is selected
\begin{align}
  \arg\max_{a} \mathbb{H}[\bar{y}_{a}|\vec{X}_{a},\mathcal{D}],
  \label{eq:entropy-based}
\end{align}
where $\mathbb{H}[x]=-\int p(x)\log p(x)dx$ represents the entropy,
and $\mathcal{D}=\{(\vec{X}_{a},\bar{y}_{a})\}_{a=1}^{A}$
is the current set of labeled sets with the aggregated outputs.

With linear basis function models in Eq.~(\ref{eq:linear}),
the entropy in Eq.~(\ref{eq:entropy-based}) can be calculated in a closed form.
We assumes the following Gaussian distribution
for the prior of parameters $\vec{w}$,
\begin{align}
  p(\vec{w})=\mathcal{N}(\vec{w}|\bm{0},\lambda^{-1}\vec{I}),
  \label{eq:prior}
\end{align}
where $\lambda\in\mathbb{R}_{>0}$ is the precision hyperparameter.

\begin{figure}
  \centering
  \includegraphics[width=15em]{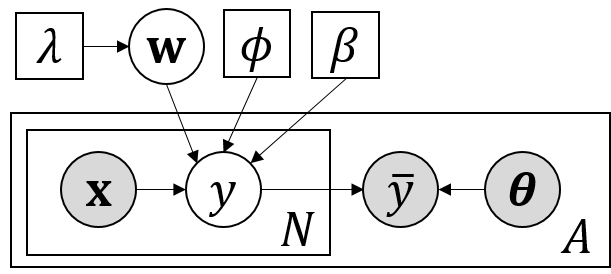}
  \caption{Graphical model representation of the proposed model. The shaded and unshaded circles represent observed and unobserved variables, respectively, the squares represent hyperparameters that can be trained, the rectangles represent repetition, and the numbers at the corner represent the number of repetitions.}
  \label{fig:graphical}
\end{figure}

We assume the following Gaussian observation noise
for the non-aggregated output given an input vector,
\begin{align}
  p(y|\vec{x},\vec{w})=\mathcal{N}(y|\hat{y}(\vec{x};\vec{w}),\beta^{-1}),
  \label{eq:py}
\end{align}
where $\beta\in\mathbb{R}_{>0}$ is the precision hyperparameter.
Then,
the distribution of aggregated output $\bar{y}_{a}$
given set of input vectors $\vec{X}_{a}$ and parameters $\vec{w}$
is the following Gaussian distribution using Eqs.~(\ref{eq:sum_of_g},\ref{eq:bary}),
\begin{align}
  p(\bar{y}_{a}|\vec{X}_{a},\vec{w})=\mathcal{N}\left(\bar{y}_{a}|\bm{\theta}_{a}^{\top}\hat{\vec{y}}_{a},\beta^{-1}\parallel\bm{\theta}_{a}\parallel^{2}\right),
  \label{eq:pya}
\end{align}
where $\hat{\vec{y}}_{a}=[\hat{y}(\vec{x}_{a1};\vec{w}),\cdots,\hat{y}(\vec{x}_{aN_{a}};\vec{w})]$,
Note that output values $y_{a1},\cdots,y_{aN_{a}}$ to be aggregated are
independent Gaussian random variables given parameters $\vec{w}$
as described in Eq.~(\ref{eq:py}).
The likelihood of labeled set $\mathcal{D}$ is given by
\begin{align}
p(\mathcal{D}|\vec{w})=\prod_{a=1}^{N_{a}}p(\bar{y}_{a}|\vec{X}_{a},\vec{w}).
\label{eq:likelihood}
\end{align}
Figure~\ref{fig:graphical} shows the graphical model representation of the proposed model.

Using the Bayes rule and Eqs.~(\ref{eq:linear},\ref{eq:prior},\ref{eq:pya},\ref{eq:likelihood}),
the posterior distribution of parameters $\vec{w}$ given $\mathcal{D}$
is given by
\begin{align}
  p(\vec{w}|\mathcal{D})=\frac{p(\vec{w})p(\mathcal{D}|\vec{w})}{\int p(\vec{w})p(\mathcal{D}|\vec{w})d\vec{w}}=
  \mathcal{N}(\vec{w}|\vec{m},\vec{S}),  
  \label{eq:posterior}
\end{align}
where
\begin{align}
  \vec{m}
  =\beta\vec{S}\sum_{a=1}^{A}\frac{\bar{y}_{a}}{\parallel\bm{\theta}_{a}\parallel^{2}}\bm{\Phi}_{a}\bm{\theta}_{a}\in\mathbb{R}^{K},
\end{align}
\begin{align}
  \vec{S}^{-1}
  =
\lambda\vec{I}+\beta\sum_{a=1}^{A}\frac{1}{\parallel\bm{\theta}_{a}\parallel^{2}}\left(\bm{\Phi}_{a}\bm{\theta}_{a}\right)\left(\bm{\Phi}_{a}\bm{\theta}_{a}\right)^{\top}\in\mathbb{R}^{K\times K},
\end{align}
and $\bm{\Phi}_{a}=[\phi(\vec{x}_{a1}),\cdots,\phi(\vec{x}_{aN_{a}})]\in\mathbb{R}^{K \times N_{a}}$.
By using a conjugate Gaussian prior in Eq.~(\ref{eq:prior}) with
 a Gaussian observation noise in Eq.~(\ref{eq:py}), the posterior is Gaussian,
and its mean and covariance are calculated in a closed form.

The joint predictive distribution of output vectors $\vec{y}_{a}=[y_{a1},\cdots,y_{aN_{a}}]\in\mathbb{R}^{N_{a}}$
for input vectors $\vec{X}_{a}$ is given by
\begin{align}
  p(\vec{y}_{a}|\vec{X}_{a},\mathcal{D})&=\int \prod_{n=1}^{N_{a}}p(y_{an}|\vec{x}_{an},\vec{w})p(\vec{w}|\mathcal{D})d\vec{w}
  =
  \mathcal{N}(\vec{y}_{a}|\vec{m}^{\top}\bm{\Phi}_{a},\beta^{-1}\vec{I}+\bm{\Phi}_{a}^{\top}\vec{S}\bm{\Phi}_{a}),
  \label{eq:joint}
\end{align}
using Eqs.~(\ref{eq:linear},\ref{eq:py},\ref{eq:posterior}).
Then, the predictive distribution of aggregated output
$\bar{y}_{a}$ is 
\begin{align}
p(\bar{y}_{a}|\vec{X}_{a},\mathcal{D})
=\mathcal{N}(\bar{y}_{a}|\vec{m}^{\top}\bm{\Phi}_{a}\bm{\theta}_{a},
\bm{\theta}_{a}^{\top}(\beta^{-1}\vec{I}+\bm{\Phi}_{a}^{\top}\vec{S}\bm{\Phi}_{a})\bm{\theta}_{a}),
\label{eq:predictive}
\end{align}
using Eqs.~(\ref{eq:sum_of_g},\ref{eq:bary},\ref{eq:joint}).

Since the entropy of the Gaussian distribution is 
\begin{align}
  \mathbb{H}[\mathcal{N}(\mu,\sigma^{2})]=\frac{1}{2}(\log\sigma^{2}+\log(2\pi)+1),
  \label{eq:gaussian_entropy}
\end{align}
the entropy of the aggregated output
in Eq.~(\ref{eq:entropy-based}) is given by
\begin{align}
  \mathbb{H}[\bar{y}_{a}|\vec{X}_{a},\mathcal{D}]&=
  \frac{1}{2}(\log
  \bm{\theta}_{a}^{\top}(\beta^{-1}\vec{I}
  +\bm{\Phi}_{a}^{\top}\vec{S}\bm{\Phi}_{a})\bm{\theta}_{a}
  +\log(2\pi)+1),
  \label{eq:entropy-based0}
\end{align}
using Eqs.~(\ref{eq:predictive},\ref{eq:gaussian_entropy}).
Intuitively speaking,
when a set contains input vectors whose
predicted outputs $\vec{y}_{a}$ are highly correlated
and have high variance,
the entropy becomes high, and such set is likely to be selected.

\subsection{Mutual information-based active learning}
\label{sec:mutual}

In the non-aggregated setting,
the mutual information-based active learning selects an instance
that maximizes the mutual information between
the output and parameters~\cite{mackay1992information,lawrence2003fast,krishnapuram2004semi}.
In the aggregated setting,
a set that maximizes the mutual information between 
aggregated output $\bar{y}_{a}$ and parameters $\vec{w}$
are selected,
\begin{align}
\arg\max_{a}
\mathbb{I}[\bar{y}_{a},\vec{w}|\mathcal{D}],
\label{eq:mi-based}
\end{align}
where $\mathbb{I}[x,x']=\int p(x,x')\log\frac{p(x,x')}{p(x)p(x')}dxdx'$
represents the mutual information,
and $\mathcal{D}$ is the current labeled aggregated data.

The mutual information in Eq.~(\ref{eq:mi-based}) equals to
the decrease of the uncertainty of the
parameters by observing the aggregated output,
\begin{align}
\mathbb{I}[\bar{y}_{a},\vec{w}|\mathcal{D}]=
  \mathbb{H}[\vec{w}|\mathcal{D}]-
  \mathbb{E}_{p(\bar{y}_{a}|\vec{X}_{a},\mathcal{D})}[\mathbb{H}[\vec{w}|\bar{y}_{a},\vec{X}_{a},\mathcal{D}]],
  \label{eq:mi-based2}
\end{align}
where
$\mathbb{E}_{p(\bar{y}_{a}|\vec{X}_{a},\mathcal{D})}[\cdot]=\int p(\bar{y}_{a}|\vec{X}_{a},\mathcal{D})[\cdot]d\vec{w}$
represents the expectation,
the first term is the entropy of the parameters given the current labeled aggregated data,
and the second term is the expected entropy of the parameter
when aggregated output $\bar{y}_{a}$ is additionally observed.
Since aggregated output $\bar{y}_{a}$ has not been observed yet,
the expectation is taken in the second term using current
model $p(\bar{y}_{a}|\vec{X}_{a},\mathcal{D})$.
We can rearrange the mutual information as follows~\cite{houlsby2011bayesian,iwata2013active,gal2017deep},
\begin{align}
\mathbb{I}[\bar{y}_{a},\vec{w}|\mathcal{D}]=
  \mathbb{H}[\bar{y}_{a}|\vec{X}_{a},\mathcal{D}]-
  \mathbb{E}_{p(\vec{w}|\mathcal{D})}[\mathbb{H}[\bar{y}_{a}|\vec{X}_{a},\vec{w}]],
  \label{eq:mi}
\end{align}
since the mutual information is symmetric,
$\mathbb{I}[\bar{y}_{a},\vec{w}|\mathcal{D}]
=\mathbb{I}[\vec{w},\bar{y}_{a}|\mathcal{D}]$.
The first term is equivalent to the objective function of the entropy-based active learning in Eq.~(\ref{eq:entropy-based}).
The entropy in the second term
is calculated by
\begin{align}
  \mathbb{H}[y_{a}|\vec{X}_{a},\vec{w}]=\frac{1}{2}(\log \beta^{-1}\parallel\bm{\theta}_{a}\parallel^{2}+\log(2\pi)+1),
  \label{eq:h2}
\end{align}
using Eqs.~(\ref{eq:pya},\ref{eq:gaussian_entropy}).
Then, the mutual information for linear basis function models is given by
\begin{align}
\mathbb{I}[\bar{y}_{a},\vec{w}|\mathcal{D}]&=
  \frac{1}{2}(\log
  \bm{\theta}_{a}^{\top}(\beta^{-1}\vec{I}+\bm{\Phi}_{a}^{\top}\vec{S}\bm{\Phi}_{a})\bm{\theta}_{a}
  -\log \beta^{-1}\parallel\bm{\theta}_{a}\parallel^{2}),
  \label{eq:mi-based0}  
\end{align}
using Eqs.~(\ref{eq:entropy-based0},\ref{eq:mi},\ref{eq:h2}).


\subsection{Procedures}
\label{sec:procedures}

Algorithm~\ref{alg} shows the active learning procedures by the proposed method.
In Line~8,
we estimate precision hyperparameters $\beta$ and $\lambda$ by
maximizing the following log marginal likelihood,
\begin{align}
  \log p(\mathcal{D}|\beta,\lambda) &= \log \int p(\mathcal{D}|\vec{w})p(\vec{w}) d\vec{w} \nonumber\\
  &=
  \frac{A}{2}\log \beta + \frac{K}{2}\log \lambda -
  \frac{1}{2}\sum_{a=1}^{A}\log \parallel\bm{\theta}_{a}\parallel^{2}
  +\frac{1}{2}\log|\vec{S}|
  -\frac{1}{2}\beta\sum_{a=1}^{A}\frac{\bar{y}_{a}^{2}}
  {\parallel\bm{\theta}_{a}\parallel^{2}}
  +\frac{1}{2}\vec{m}^{\top}\vec{S}^{-1}\vec{m}
  -\frac{A}{2}\log(2\pi),
  \label{eq:marginal}
\end{align}
which is derived using Eqs.~(\ref{eq:prior},\ref{eq:likelihood}),
where we omit the precision hyperparameters
in $p(\mathcal{D}|\vec{w},\beta)=p(\mathcal{D}|\vec{w})$
and $p(\vec{w}|\lambda)=p(\vec{w})$ for simplicity.
When we use neural networks for basis function $\phi(\vec{x})$,
we can also estimate the neural network parameters
by maximizing this log marginal likelihood.
The predictive distribution of the non-aggregated output given an input vector
is calculated by
\begin{align}
p(y|\vec{x},\mathcal{D})=
\int p(y|\vec{x},\vec{w})p(\vec{w}|\mathcal{D})d\vec{w}
=
\mathcal{N}(y|\vec{m}^{\top}\phi(\vec{x}),\beta^{-1}+\phi(\vec{x})^{\top}\vec{S}\phi(\vec{x})),
\end{align}
using Eq.~(\ref{eq:py}) and the posterior in Eq.~(\ref{eq:posterior}).

\begin{algorithm}[t!]
  \caption{Active learning procedures with aggregated outputs by the proposed method.}
  \label{alg}
  \begin{algorithmic}[1]
    \renewcommand{\algorithmicrequire}{\textbf{Input:}}
    \renewcommand{\algorithmicensure}{\textbf{Output:}}
    \REQUIRE{Unlabeled sets of input vectors $\mathcal{U}=\{\vec{X}_{a}\}_{a=1}^{A}$, number of queries $T$.}
    \ENSURE{Posterior probability of model parameters $\vec{w}$.}
    \STATE Initialize precision hyperparameters $\lambda$, $\beta$ randomly. 
    \STATE Initialize labeled sets with an empty set $\mathcal{D}= \{\}$.
    \FOR{each query $t=1,\cdots,T$}
    \STATE Select set $\hat{a}$ that maximizes the entropy in Eq.~(\ref{eq:entropy-based0}) or mutual information in Eq.~(\ref{eq:mi-based0}) from unlabeled sets $\mathcal{U}$.
    \STATE Observe aggregated output $\bar{y}_{\hat{a}}$ of the selected set.
    \STATE Include selected set $\hat{a}$ into training data $\mathcal{D}=\mathcal{D}\cup(\vec{X}_{\hat{a}},y_{\hat{a}})$.
    \STATE Exclude selected set $\hat{a}$ from unlabeled sets $\mathcal{U}=\mathcal{U}\setminus \vec{X}_{\hat{a}}$.
    \STATE Update precision hyperparameters $\lambda, \beta$ by maximizing the log marginal likelihood in Eq.~(\ref{eq:marginal}).
    \STATE Estimate posterior $p(\vec{w}|\mathcal{D})$ by Eq.~(\ref{eq:posterior}).
    \ENDFOR
  \end{algorithmic}
\end{algorithm}


\section{Experiments}
\label{sec:experiments}
\subsection{Data}

We evaluated the proposed method using regression datasets in LIBSVM
~\cite{chang2011libsvm}~\footnote{The LIBSVM datasets were obtained
from~\url{https://www.csie.ntu.edu.tw/~cjlin/libsvmtools/datasets/}.},
and the climate data in North American~\footnote{The climate datasets were obtained from \url{https://sites.ualberta.ca/~ahamann/data/climatena.html}.}.

Table~\ref{tab:libsvm} shows the statistics of the LIBSVM datasets.
For each of the LIBSVM datasets,
we randomly selected 80\% of the instances for training,
and the remaining for testing.
Using the training instances, we generated sets of aggregated data
by randomly selecting instances without replacement,
where the number of instances in a set is between one to 20.

The climate data contain 26 bio-climate values,
such as mean annual temperature,
length of the frost-free period,
precipitation as snow,
summer heat-moisture index,
and annual heat-moisture index,
for each location defined by longitude, latitude, and elevation.
One of the 26 bio-climate values was used as outputs,
and the others including location information were used as attributes.
In total, 26 datasets for regression with 28 attributes and a scalar output
were generated.
We subsampled 336 locations (instances) with the interval of
three degrees of longitude and latitude.
The data were aggregated by the state,
and 72 sets were obtained.
The minimum, average, and maximum numbers of instances in a set were 1,
4.67, and 51.
We randomly selected 80\% of the sets for training,
and instances in the remaining sets were used for testing.

For each set in both the LIBSVM and climate datasets,
the aggregated output was calculated by
summing the individual outputs in each set.
Note that when the aggregated output is the average of the outputs,
we can transform it to the sum of the outputs
by multiplying the number of instances without knowing individual outputs
as preprocessing.
The test data were not aggregated,
where the output for each test instance was predicted.
The number of queries to observe aggregated outputs was $T=30$.
We performed 50 experiments with different splits of training and test data
for each dataset.
We did not use validation data since
we considered a situation where
a limited number of labeled sets are available.

\begin{table}[t!]
\centering
\caption{Statistics of LIBSVM datasets: the number of instances, and the number of attributes $D$.}
\label{tab:libsvm}
\begin{tabular}{lrrr}
\hline
& \#instances & \#attributes \\
\hline
Abalone & 4,177 & 8\\
Cadata & 20,640 & 8\\
Cpu & 8,192 & 12\\
Housing & 506 & 13\\
Mg & 1,385 & 6\\
Space & 3,107 & 6\\
\hline
\end{tabular}
\end{table}

\subsection{Comparing methods}

We compared the proposed active learning methods for aggregated data
based on the mutual information (AggMI) and entropy (AggEnt)
with the following methods: the mutual information-based method for non-aggregated data (MI),
the entropy-based method for non-aggregated data (Ent),
the variance of the input vectors (Var),
query by committee~\cite{burbidge2007active} (QBC),
expected model change maximization~\cite{cai2013maximizing} (EMCM),
the maximum of the number of instances (MaxN),
the minimum of the number of instances (MinN),
and random (Rand).

MI (Ent) selects a set that maximizes
the sum of the mutual informations (entropies)
over the instances in the set.
MI (Ent) corresponds to AggMI (AggEnt)
without consideration of covariance across instances in a set,
and can be seen as existing mutual information-based (entropy-based)
active learning method for non-aggregated data.
Var selects a set whose variance in the input vectors is maximum.
QBC selects a set whose variance in the predictions
of the committee members is maximum.
The predictions of the committee members were obtained by
the models with parameters that are sampled from the posterior.
EMCM selects a set that maximizes the expected model change,
which was approximated with an ensemble that was created as in QBC.
MaxN (MinN) selects a set that maximizes (minimizes)
the number of instances.
Rand randomly selects a set.

For all methods including the proposed method,
we used approximated Gaussian processes~\cite{rahimi2007random}
for modeling the non-aggregated output.
With the approximated Gaussian processes,
the basis function in Eq.~\ref{eq:linear} was set as follows,
\begin{align}
  \phi(\vec{x})=\left[\sqrt{\frac{2}{K-1}}\cos(-\vec{B}\vec{x}+\vec{c}), 1\right],
  \label{eq:phi}
\end{align}
where $\vec{B}\in\mathbb{R}^{(K-1)\times D}$ is generated from
the standard Gaussian distribution, $b_{kd}\sim\mathcal{N}(0,1)$,
and $\vec{c}\in\mathbb{R}^{K-1}$ is generated from
a unifrom distribution $c_{k}\sim\mathrm{Uniform}(0,2\pi)$.
In Eq.~(\ref{eq:phi}), the last element of the basis function
is set to one, $\phi_{K}(x)=1$, for a bias parameter.
We used $K=128$.
The precision hyperparameters were trained by maximizing the marginal likelihood
in Eq.~(\ref{eq:marginal}) using Adam~\cite{kingma2014adam}
with learning rate $10^{-3}$ and 1000 epochs.
We implemented the proposed method with PyTorch~\cite{NEURIPS2019_bdbca288}.

\subsection{Results}

\begin{table*}[t!]
\centering
\caption{Mean squared error averaged over different numbers of queries from one to 30. Values in bold typeface are not statistically different at 5\% level from the best performing method in each row by a paired t-test. The bottom row shows the number of datasets the method achieved the performance that are not statistically different from the best performing method.}
\label{tab:mse}
{\tabcolsep=0.8em\begin{tabular}{lrrrrrrrrrr}
\hline
& AggMI & AggEnt & MI & Ent & QBC & EMCM & Var & MaxN & MinN & Rand \\
\hline
Abalone & {\bf 0.041} & {\bf 0.042} & 0.052 & 0.048 & 0.044 & 0.046 & 0.056 & 0.048 & 0.059 & 0.051\\
Cadata & {\bf 0.183} & {\bf 0.185} & 0.228 & 0.206 & 0.199 & 0.218 & 0.238 & 0.227 & 0.353 & 0.221\\
Cpu & {\bf 0.031} & 0.033 & 0.168 & 0.043 & 0.036 & 0.044 & 0.125 & 0.043 & 0.104 & 0.046\\
Housing & {\bf 0.119} & 0.124 & 0.133 & 0.129 & 0.124 & 0.128 & 0.137 & 0.130 & 0.144 & 0.128\\
Mg & {\bf 0.199} & 0.209 & 0.234 & 0.213 & 0.214 & 0.215 & 0.225 & 0.211 & 0.234 & 0.226\\
Space & {\bf 0.011} & {\bf 0.011} & 0.015 & 0.012 & 0.011 & 0.013 & 0.014 & 0.013 & 0.017 & 0.013\\
eFFP & {\bf 0.080} & 0.093 & 0.147 & 0.089 & 0.095 & 0.097 & 0.167 & 0.112 & 0.387 & 0.157\\
bFFP & {\bf 0.081} & 0.089 & 0.150 & 0.088 & 0.090 & 0.090 & 0.119 & 0.100 & 0.394 & 0.137\\
Tave-wt & {\bf 0.081} & 0.090 & 0.158 & 0.086 & 0.092 & 0.092 & 0.170 & 0.105 & 0.473 & 0.171\\
Tave-sm & {\bf 0.094} & {\bf 0.099} & 0.119 & {\bf 0.093} & {\bf 0.092} & {\bf 0.091} & 0.136 & 0.103 & 0.207 & {\bf 0.108}\\
TD & {\bf 0.072} & 0.076 & 0.107 & {\bf 0.074} & 0.076 & 0.077 & 0.180 & 0.087 & 0.524 & 0.180\\
SHM & {\bf 0.028} & 0.030 & 0.040 & 0.032 & 0.029 & 0.031 & 0.079 & 0.031 & 0.039 & 0.041\\
RH & {\bf 0.106} & {\bf 0.105} & 0.143 & 0.107 & 0.108 & {\bf 0.106} & 0.132 & 0.112 & 0.130 & 0.133\\
PPT-wt & {\bf 0.052} & 0.053 & 0.066 & 0.053 & 0.053 & 0.054 & 0.078 & 0.056 & 0.099 & 0.066\\
PPT-sm & {\bf 0.045} & 0.053 & 0.194 & 0.052 & 0.053 & 0.055 & 0.064 & 0.062 & 0.273 & 0.084\\
PAS & {\bf 0.024} & 0.028 & 0.029 & 0.027 & 0.027 & 0.028 & 0.029 & 0.031 & 0.036 & 0.030\\
NFFD & {\bf 0.125} & 0.145 & 0.248 & 0.139 & 0.144 & 0.148 & 0.212 & 0.170 & 0.550 & 0.232\\
MWMT & {\bf 0.061} & {\bf 0.063} & 0.097 & {\bf 0.063} & {\bf 0.063} & {\bf 0.064} & 0.113 & 0.071 & 0.272 & 0.115\\
MSP & {\bf 0.053} & 0.060 & 0.171 & 0.059 & 0.061 & 0.063 & 0.080 & 0.069 & 0.285 & 0.098\\
MCMT & {\bf 0.068} & 0.078 & 0.137 & 0.075 & 0.078 & 0.079 & 0.175 & 0.092 & 0.481 & 0.169\\
MAT & {\bf 0.082} & 0.088 & {\bf 0.094} & 0.089 & {\bf 0.087} & 0.090 & 0.133 & 0.104 & 0.338 & 0.116\\
MAP & {\bf 0.040} & 0.041 & 0.063 & 0.042 & {\bf 0.041} & 0.042 & 0.054 & 0.043 & 0.158 & 0.070\\
FFP & {\bf 0.088} & 0.101 & 0.236 & 0.096 & 0.102 & 0.104 & 0.123 & 0.125 & 0.451 & 0.162\\
Eref & {\bf 0.102} & 0.117 & 0.327 & 0.111 & 0.118 & 0.120 & 0.154 & 0.143 & 0.487 & 0.227\\
EXT & {\bf 0.075} & {\bf 0.074} & 0.099 & {\bf 0.074} & {\bf 0.074} & {\bf 0.074} & 0.125 & 0.079 & 0.190 & 0.127\\
EMT & {\bf 0.073} & 0.081 & 0.190 & 0.078 & 0.082 & 0.083 & 0.087 & 0.094 & 0.339 & 0.136\\
DD-18 & 0.121 & 0.128 & {\bf 0.099} & 0.118 & 0.125 & 0.126 & {\bf 0.112} & 0.139 & 0.383 & 0.190\\
DD-0 & 0.329 & 0.327 & {\bf 0.092} & 0.329 & 0.326 & 0.328 & 0.111 & 0.328 & 0.297 & 0.172\\
DD5 & {\bf 0.057} & 0.068 & 0.130 & 0.065 & 0.068 & 0.071 & 0.161 & 0.084 & 0.477 & 0.194\\
DD18 & {\bf 0.047} & 0.054 & 0.337 & 0.051 & 0.054 & 0.055 & 0.115 & 0.066 & 0.457 & 0.129\\
CMD & {\bf 0.085} & 0.095 & {\bf 0.089} & 0.091 & 0.095 & 0.097 & 0.147 & 0.108 & 0.353 & 0.175\\
AHM & {\bf 0.018} & {\bf 0.018} & 0.024 & 0.018 & 0.018 & 0.018 & 0.036 & 0.019 & 0.042 & 0.030\\
\hline
\# best & 30 & 8 & 4 & 4 & 5 & 4 & 1 & 0 & 0 & 1 \\
\hline
\end{tabular}}
\end{table*}

\begin{figure*}[t!]
\centering
{\tabcolsep=0.6em
\begin{tabular}{ccc}
\includegraphics[width=13.8em]{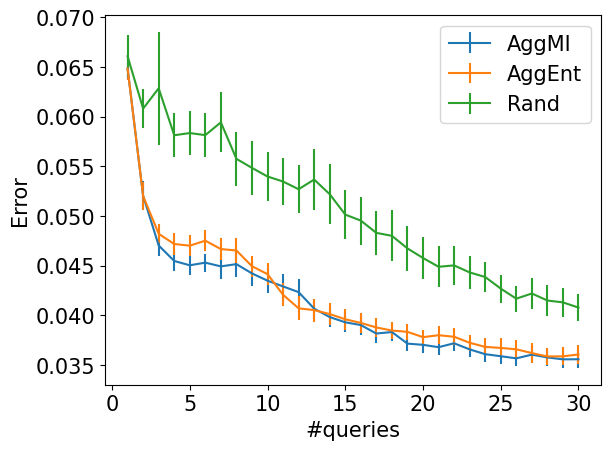}&
\includegraphics[width=13.8em]{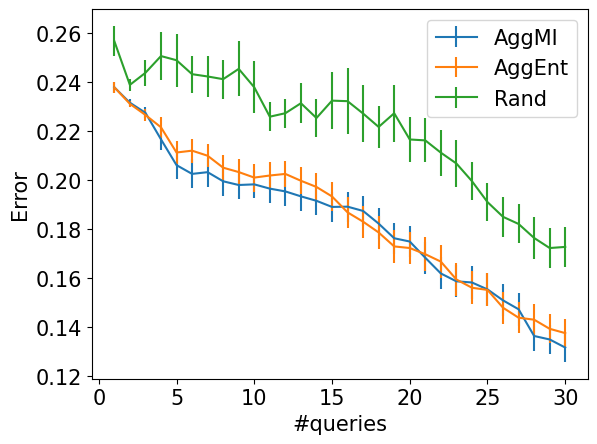}&
\includegraphics[width=13.8em]{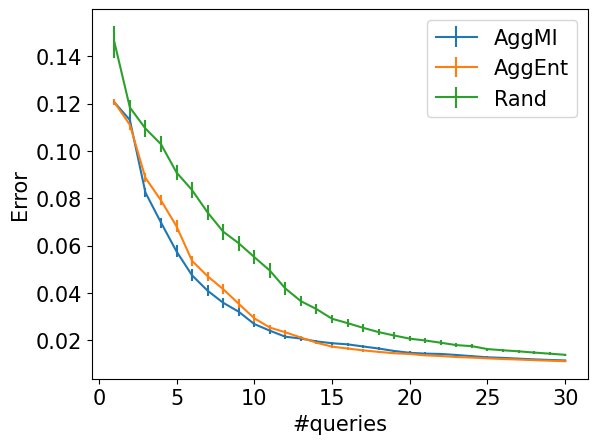}\\
(a) Abalone & (b) Cadata & (c) Cpu \\
\includegraphics[width=13.8em]{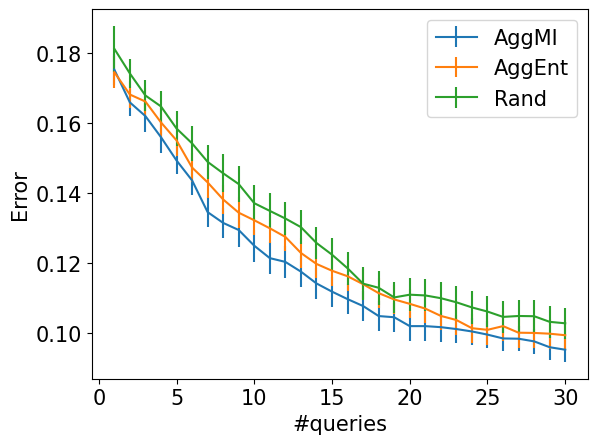}&
\includegraphics[width=13.8em]{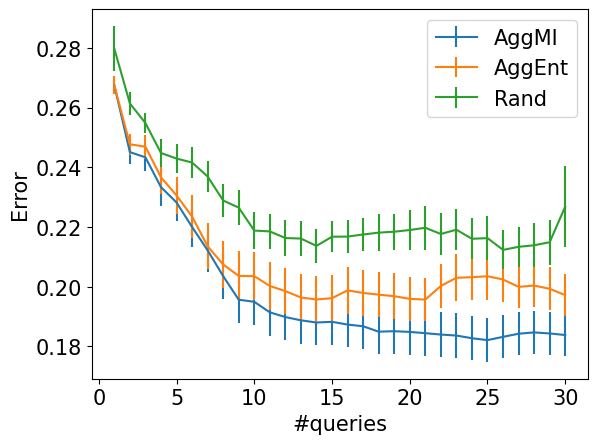}&
\includegraphics[width=13.8em]{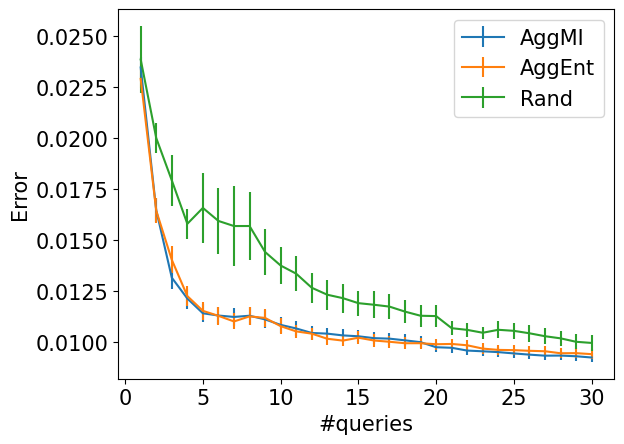}\\
(d) Housing & (e) Mg & (g) Space \\
\includegraphics[width=13.8em]{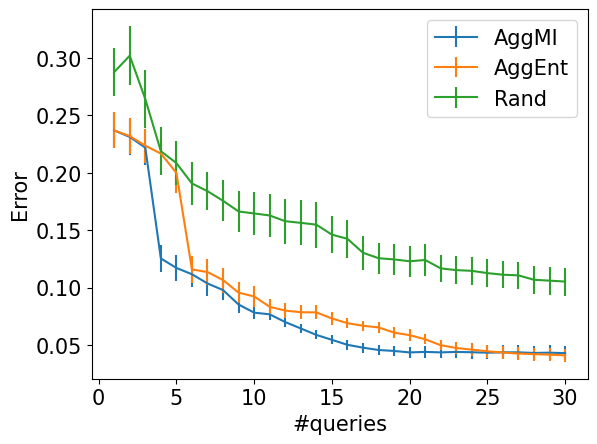}&
\includegraphics[width=13.8em]{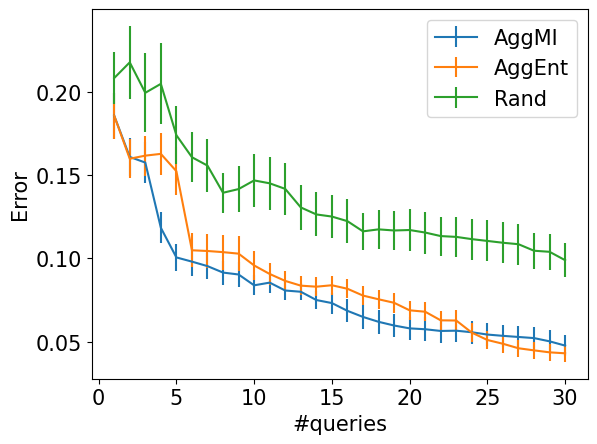}&
\includegraphics[width=13.8em]{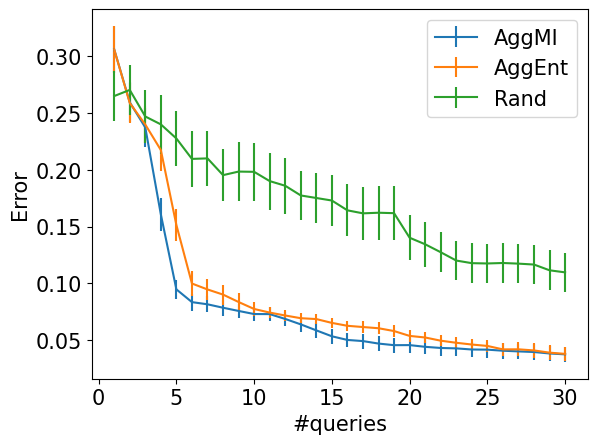}\\
(h) eFFP & (i) bFFP & (j) Tave-wt \\
\end{tabular}}
\caption{Mean squared error with different numbers of queries. Bars shows the standard error.}
\label{fig:libsvm}
\end{figure*}

\begin{figure*}[t!]
\centering
{\tabcolsep=0.6em
\begin{tabular}{ccc}
\includegraphics[width=13.8em]{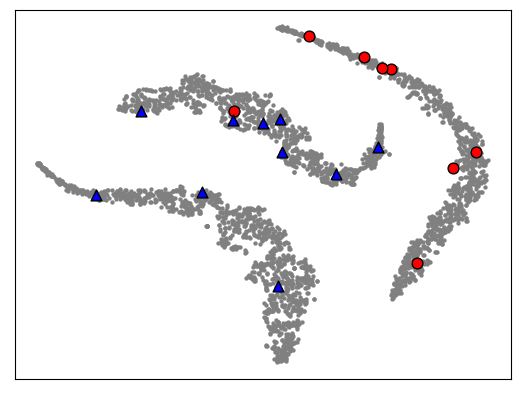} &
\includegraphics[width=13.8em]{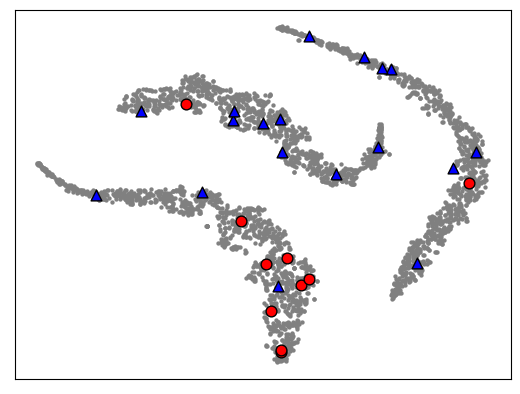} &
\includegraphics[width=13.8em]{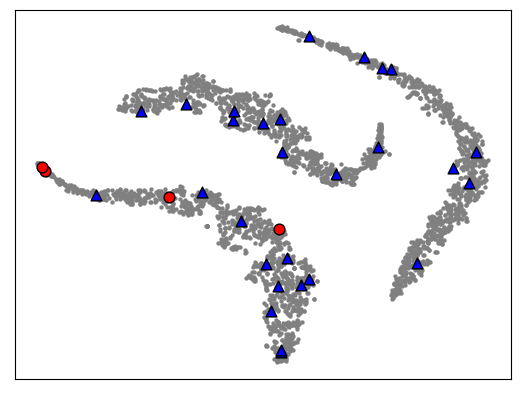} \\
(a) 2nd query & (b) 3rd query & (c) 4th query \\
\end{tabular}}
\caption{Visualization by t-SNE~\cite{van2008visualizing} of the Abalone dataset with different numbers of queries. Red circles are the instances in the newly selected set, blue triangles are the instances in the already selected sets, and gray dots are unlabeled instances.}
\label{fig:sne}
\end{figure*}

\begin{table*}[t!]
\centering
\caption{Training computational time in seconds with the Abalone dataset.}
\label{tab:time}
{\tabcolsep=0.5em\begin{tabular}{rrrrrrrrrr}
\hline
AggMI & AggEnt & MI & Ent & QBC & EMCM & Var & MaxN & MinN & Rand \\
\hline
2599.831 & 2544.971 & 2521.719 & 2577.427 & 3072.736 & 3111.591 & 3036.683 & 2542.546 & 2425.869 & 2523.622\\
\hline
\end{tabular}}
\end{table*}

Table~\ref{tab:mse} shows the mean squared error averaged over different numbers of queries.
The proposed AggMI method achieved the lowest mean squared error on most datasets.
The AggEnt method selects sets whose labels are the most uncertain.
Therefore, although the AggEnt method can reduce the uncertainty in the selected sets,
it does not necessarily reduce the uncertainty of the model parameters.
On the other hand, since the AggMI method selects sets that directly maximize
the decrease of the uncertainty of the parameters as in Eq.~(\ref{eq:mi-based2}),
it performed better than the AggEnt method.
The methods that do not consider covariance in each set (MI and Ent)
performed worse than those that consider covariance (AggMI and AggEnt).
This result indicates that the covariance is important for active learning with aggregated outputs.
The active learning methods
for non-aggregated data (QBC and EMCM) did not perform well.
The variance of the input vectors (Var) was worse than AggEnt,
which considers the variance of the aggregated outputs.
The error by the MaxN method was lower than that by the MinN method.
Figure~\ref{fig:libsvm} shows the mean squared error for each number of queries.
All methods decreased the error as the number of queries with most datasets.

Figure~\ref{fig:sne} shows the visualization of the Abalone dataset
at different numbers of queries by the AggMI method,
where the aggregated data were generated by randomly
selected sets with sizes one to ten.
The data consisted of three clusters.
For each query, the AggMI method selected sets that comprised instances in a certain cluster
that contained fewer labeled instances.
For example, at the 2nd query, a set that contained seven instances in the right cluster was selected,
where there have been no labeled instances in the right cluster.
It is because observing labels of such sets is likely to decrease
the uncertainty of the model parameters.
Also, the AggMI method selected sets where most of the instances were in the same cluster.
For example, seven out of eight instances were in the right cluster in the selected set at the 2nd query.
It is because the AggMI method favors sets with highly correlated instances since the aggregated output of such sets is likely to have high entropy.

Table~\ref{tab:time} shows the training computational time in seconds
on the Abalone dataset using computers with 2.60GHz CPUs.
The computational time was almost the same among all methods.
In the training phase, estimating hyperparameters is time-consuming.
The calculation of the entropy and mutual information with the proposed method
is efficient since they can be obtained in a closed form as in Eqs.~(\ref{eq:entropy-based0},\ref{eq:mi-based0}).


\subsection{Results with neural network models}

We evaluated the proposed method with neural networks for modeling the non-aggregated output.
For all methods including the proposed method,
the basis function is modeled by four-layered feed-forward neural networks
with 32 hidden units, and 128 output units.
The neural network parameters were trained by maximizing the marginal likelihood using Adam~\cite{kingma2014adam}
with learning rate $10^{-3}$ and 1000 epochs.
In addition to the comparing methods described in the paper,
we compared with dropout-based active learning methods: dropout-based AggMI (DAMI)
and dropout-based AggEnt (DAEnt). With the dropout-based methods,
the entropy and mutual information
are calculated by the Monte Carlo method with dropout~\cite{srivastava2014dropout,gal2017deep}. 
By approximating the predictive distribution of aggregated output $\bar{y}_{a}$
with a Gaussian distribution, the variance is given by
\begin{align}
  \sigma^{2}&=\frac{1}{L}\sum_{\ell=1}^{L}\left(\sum_{n=1}^{N_{a}}\theta_{n}\hat{y}'(\vec{x}_{an};\vec{w}_{\ell})\right)^{2}
  -\left(\frac{1}{L}\sum_{\ell=1}^{L}\sum_{n=1}^{N_{a}}\theta_{n}\hat{y}'(\vec{x}_{an};\vec{w}_{\ell})\right)^{2},  
\end{align}
where $L$ is the number of dropout samples, and
$\vec{w}_{\ell}$ is the $\ell$th sampled parameters by dropout,
and $\hat{y}'(\vec{x}_{an};\vec{w}_{\ell})=\hat{y}(\vec{x}_{an};\vec{w}_{\ell})+\beta\epsilon$
is the non-aggregated output prediction by a neural network with parameter $\vec{w}_{\ell}$
and standard Gaussian noise $\epsilon$.
The entropy of the aggregated output (DAEnt) is given by
\begin{align}
  \mathbb{H}[\bar{y}_{a}|\vec{X}_{a},\mathcal{D}]=\frac{1}{2}(\log\sigma^{2}+\log(2\pi)+1).
  \label{eq:entropy-based0_NN}  
\end{align}
Intuitively speaking,
sets where the aggregated output predictions vary across different dropout samples are likely to be selected.
The mutual information (DAMI) is calculated by
\begin{align}
\mathbb{I}[\bar{y}_{a},\vec{w}|\mathcal{D}]&=
\frac{1}{2}(\log \sigma^{2}-\log\beta^{-1}\parallel\bm{\theta}_{a}\parallel^{2}).
  \label{eq:mi-based0_NN}  
\end{align}
The results are shown in Table~\ref{tab:msenn}.
The proposed method (AggMI and AggEnt) achieved the better performance than the other methods.
Since the neural network-based basis functions have many parameters that are not considered as random variables,
the estimated entropy and mutual information were not accurate
compared with those with the approximated Gaussian processes by random features.
Therefore, the difference of the performance between the proposed method and the other methods,
with the neural network-based basis functions in Table~\ref{tab:msenn}
were smaller than that with the approximated Gaussian processes in the main paper.
For the same reason, the errors by AggMI and AggEnt were similar.
Dropout-based methods (DAMI and DAEnt) were worse than the proposed method.
This result indicates the effectiveness of the Bayesian linear basis function models.

\begin{table*}[t!]
\centering
\caption{Mean squared error averaged over different numbers of queries from one to 30. Values in bold typeface are not statistically different at 5\% level from the best performing method in each row by a paired t-test. The bottom row shows the number of datasets the method achieved the performance that are not statistically different from the best performing method.}
\label{tab:msenn}
{\tabcolsep=0.3em\begin{tabular}{lrrrrrrrrrrrrrrr}
\hline
& AggMI & DAMI & AggEnt & DAEnt & MI & Ent & QBC & EMCM & Var & MaxN & MinN & Rand \\
\hline
Abalone & {\bf 0.052} & {\bf 0.050} & {\bf 0.051} & {\bf 0.052} & {\bf 0.053} & 0.052 & 0.054 & 0.052 & 0.065 & 0.053 & 0.073 & 0.053\\
Cadata & {\bf 0.200} & {\bf 0.198} & {\bf 0.207} & {\bf 0.205} & 0.219 & 0.211 & {\bf 0.201} & {\bf 0.206} & 0.232 & 0.229 & 0.284 & 0.246\\
Cpu & {\bf 0.047} & 0.074 & 0.051 & 0.071 & 0.143 & 0.073 & 0.059 & 0.055 & 0.181 & {\bf 0.052} & 0.107 & 0.056\\
Housing & 0.134 & {\bf 0.128} & 0.136 & {\bf 0.127} & {\bf 0.130} & 0.136 & 0.139 & 0.142 & 0.138 & 0.135 & {\bf 0.131} & 0.137\\
Mg & 0.206 & {\bf 0.203} & 0.219 & {\bf 0.197} & 0.219 & 0.220 & 0.218 & 0.215 & 0.233 & 0.221 & 0.224 & {\bf 0.204}\\
Space & 0.016 & {\bf 0.015} & 0.017 & {\bf 0.014} & 0.024 & 0.016 & 0.016 & 0.016 & 0.023 & 0.016 & 0.027 & 0.017\\
eFFP & 0.045 & 0.051 & {\bf 0.032} & 0.060 & 0.047 & 0.034 & 0.040 & 0.054 & 0.051 & 0.050 & 0.101 & 0.046\\
bFFP & {\bf 0.032} & 0.050 & {\bf 0.030} & 0.058 & 0.046 & 0.032 & 0.033 & 0.039 & 0.041 & 0.053 & 0.079 & 0.043\\
Tave-wt & 0.032 & 0.056 & 0.030 & 0.036 & {\bf 0.017} & 0.031 & 0.035 & 0.040 & 0.040 & 0.043 & 0.131 & 0.045\\
Tave-sm & 0.040 & 0.053 & 0.049 & 0.042 & {\bf 0.037} & 0.051 & 0.050 & 0.052 & {\bf 0.035} & 0.072 & 0.099 & {\bf 0.037}\\
TD & {\bf 0.045} & 0.083 & 0.049 & 0.060 & 0.055 & 0.051 & {\bf 0.045} & {\bf 0.047} & 0.086 & 0.059 & 0.181 & 0.072\\
SHM & 0.034 & {\bf 0.020} & {\bf 0.023} & 0.032 & 0.046 & {\bf 0.017} & 0.026 & 0.037 & 0.176 & {\bf 0.017} & 0.042 & 0.043\\
RH & {\bf 0.075} & 0.127 & {\bf 0.080} & {\bf 0.079} & {\bf 0.075} & {\bf 0.076} & 0.082 & {\bf 0.077} & {\bf 0.078} & 0.094 & 0.315 & 0.100\\
PPT-wt & {\bf 0.042} & 0.055 & {\bf 0.044} & 0.066 & 0.060 & 0.056 & 0.050 & 0.052 & 0.087 & 0.106 & 0.106 & 0.061\\
PPT-sm & {\bf 0.053} & {\bf 0.054} & 0.061 & 0.065 & {\bf 0.049} & 0.061 & 0.058 & 0.057 & 0.056 & 0.060 & 0.086 & {\bf 0.054}\\
PAS & {\bf 0.017} & 0.025 & 0.019 & 0.042 & 0.045 & 0.020 & 0.020 & 0.020 & 0.040 & 0.028 & 0.052 & 0.034\\
NFFD & {\bf 0.059} & {\bf 0.061} & {\bf 0.055} & 0.073 & {\bf 0.057} & 0.057 & 0.095 & 0.099 & {\bf 0.057} & 0.082 & 0.096 & {\bf 0.066}\\
MWMT & 0.043 & 0.056 & {\bf 0.036} & {\bf 0.038} & 0.050 & 0.042 & 0.042 & 0.041 & 0.046 & 0.042 & 0.108 & 0.049\\
MSP & {\bf 0.048} & 0.058 & 0.057 & 0.058 & {\bf 0.050} & 0.058 & 0.061 & 0.061 & 0.066 & 0.059 & 0.095 & 0.055\\
MCMT & 0.031 & 0.052 & 0.033 & 0.043 & {\bf 0.020} & 0.034 & 0.045 & 0.051 & 0.039 & 0.046 & 0.122 & 0.044\\
MAT & 0.035 & 0.047 & 0.037 & 0.041 & {\bf 0.017} & 0.045 & 0.043 & 0.045 & 0.033 & 0.050 & 0.096 & 0.037\\
MAP & {\bf 0.042} & 0.047 & {\bf 0.041} & 0.050 & 0.049 & 0.045 & 0.045 & 0.048 & 0.065 & 0.058 & 0.081 & 0.053\\
FFP & 0.051 & 0.058 & {\bf 0.043} & 0.065 & 0.057 & 0.052 & {\bf 0.050} & 0.065 & 0.056 & 0.069 & 0.108 & {\bf 0.046}\\
Eref & 0.050 & 0.055 & {\bf 0.038} & 0.060 & 0.051 & 0.052 & 0.087 & 0.094 & 0.054 & 0.069 & 0.113 & 0.051\\
EXT & 0.059 & 0.090 & {\bf 0.048} & 0.055 & 0.067 & 0.055 & 0.053 & 0.055 & {\bf 0.056} & 0.057 & 0.165 & 0.064\\
EMT & {\bf 0.034} & 0.040 & {\bf 0.034} & 0.060 & 0.049 & 0.038 & 0.044 & 0.051 & 0.050 & 0.051 & 0.102 & {\bf 0.039}\\
DD-18 & {\bf 0.042} & 0.088 & 0.049 & 0.055 & 0.073 & 0.056 & 0.060 & 0.072 & {\bf 0.048} & 0.055 & 0.187 & 0.056\\
DD-0 & 0.098 & 0.117 & {\bf 0.084} & {\bf 0.083} & {\bf 0.098} & {\bf 0.084} & {\bf 0.081} & 0.095 & {\bf 0.075} & {\bf 0.090} & 0.241 & 0.099\\
DD5 & 0.040 & 0.057 & {\bf 0.029} & 0.041 & 0.055 & 0.047 & 0.051 & 0.061 & 0.063 & 0.050 & 0.097 & 0.045\\
DD18 & {\bf 0.042} & 0.050 & 0.056 & 0.053 & 0.083 & 0.057 & 0.060 & 0.061 & 0.085 & 0.061 & 0.137 & 0.071\\
CMD & {\bf 0.063} & 0.092 & 0.076 & 0.076 & 0.150 & 0.080 & 0.085 & 0.095 & 0.110 & 0.085 & 0.143 & 0.096\\
AHM & {\bf 0.013} & 0.018 & 0.014 & 0.016 & 0.021 & 0.016 & 0.014 & 0.014 & 0.030 & {\bf 0.012} & 0.029 & 0.022\\
\hline
\# best & 17 & 8 & 16 & 8 & 11 & 3 & 4 & 3 &  6 & 4 & 1 & 6 \\
\hline
\end{tabular}}
\end{table*}

\section{Conclusion}
\label{sec:conclusion}

We proposed an active learning method for aggregated data
that sequentially selects sets to observe its aggregated output
to improve the predictive performance with fewer observations.
The proposed method selects sets using the entropy of aggregated outputs,
or mutual information between aggregated outputs and model parameters.
We derived the analytical solution of the entropy and mutual information with
Bayesian linear basis function modeling. 
We experimentally demonstrated that the proposed method achieved lower test errors than the existing active learning methods.

Although our results are encouraging, we must extend our approach in several directions.
First, we will apply our approach to non-Gaussian likelihoods for tasks such as classification
and Poisson regression using models in~\cite{bhowmik2015generalized,zhang2020learning,law2018variational}.
Second, we want to extend the proposed method such that
it selects a subset of input vectors to be labeled
from a set of all candidate input vectors
although we assume that the sets of instances are fixed and given in this paper.

\bibliographystyle{abbrv}
\bibliography{arxiv_active}

\end{document}